\documentclass{bmvc2k}
\usepackage{amsfonts}
\usepackage{booktabs}
\usepackage{multirow}
\usepackage[table]{xcolor}
\usepackage{booktabs}
\usepackage{multirow}
\usepackage{soul}
\usepackage{graphicx}
\usepackage{wrapfig}

\title{ReCalMatch: \\Reliability-Calibrated Semantic Guidance \\
for Semi-Supervised Fine-Grained Recognition}

\addauthor{Yundi Hong}{Yundi.Hong@warwick.ac.uk}{1}
\addauthor{Hongyang He$^{\dagger}$}{Hongyang.He@warwick.ac.uk}{1}
\addauthor{Zheng Fang}{Zheng.Fang.6@warwick.ac.uk}{1}
\addauthor{Xuanyu Liu}{Xuanyu.Liu@warwick.ac.uk}{1}
\addauthor{Victor Sanchez$^{\ddagger}$}{V.F.Sanchez-Silva@warwick.ac.uk}{1}

\addinstitution{
Department of Computer Science\\
University of Warwick\\
Coventry, UK
}

\runninghead{Hong, He, Fang, Liu, Sanchez}{ReCalMatch}

\begin{document}

\maketitle
{\renewcommand{\thefootnote}{\fnsymbol{footnote}}\footnotetext[2]{Corresponding author.} \footnotetext[3]{ Victor Sanchez is also with Universitat Autonoma de Barcelona, Spain, as a Distinguished Researcher. This work was partially funded by the Spanish Ministry of Science, Innovation and Universities (MICIU) -- ATR2024-154513}}
\vspace{-0.8em}

\begin{abstract}
Semi-supervised fine-grained visual recognition is highly vulnerable to overconfident pseudo-label errors: visually similar categories frequently produce high-confidence yet incorrect predictions, and consistency regularization then reinforces these errors throughout training. Existing semi-supervised learning (SSL) methods estimate pseudo-label reliability almost entirely from the visual classifier itself---maximum probability, adaptive thresholds, or entropy---signals that remain blind to whether a predicted class is \emph{semantically} compatible with the visual representation. We propose \textbf{ReCalMatch}, a reliability-calibrated semantic framework for semi-supervised fine-grained recognition. Rather than treating textual semantics as auxiliary supervision, ReCalMatch uses multi-aspect semantic prototypes as \emph{calibration evidence} for pseudo-label learning. We construct class-conditioned semantic prototypes from class names and domain-specific semantic aspects, and measure a \emph{visual--semantic agreement} score between each unlabeled embedding and its pseudo-label prototype. This agreement is combined with prediction confidence and entropy into a single reliability weight that down-weights pseudo-labels that are visually confident but semantically inconsistent. A semantic consistency term and a semantic margin regularizer further sharpen prototype separability under limited labels. Extensive experiments on CUB-200-2011, Stanford Dogs, NABirds, and iNaturalist18 show that ReCalMatch consistently improves strong SSL baselines, with the largest gains in low-label regimes where pseudo-label noise is most severe.
\end{abstract}


\section{Introduction}
Semi-supervised learning (SSL) lowers annotation costs by combining a small labeled set with a large unlabeled set, and has achieved strong results on generic image classification. Its effectiveness drops sharply, however, on fine-grained visual classification (FGVC)~\cite{wei2021fine}. Fine-grained categories are separated by subtle local cues, while intra-class appearance varies widely under pose, illumination, and background changes~\cite{krause20133d,A13}. In this regime, pseudo-labels are easily corrupted, and consistency regularization can reinforce incorrect predictions, accumulating confirmation bias as training proceeds~\cite{A7}.
\begin{figure}[t]
    \centering
    \includegraphics[width=1\linewidth]{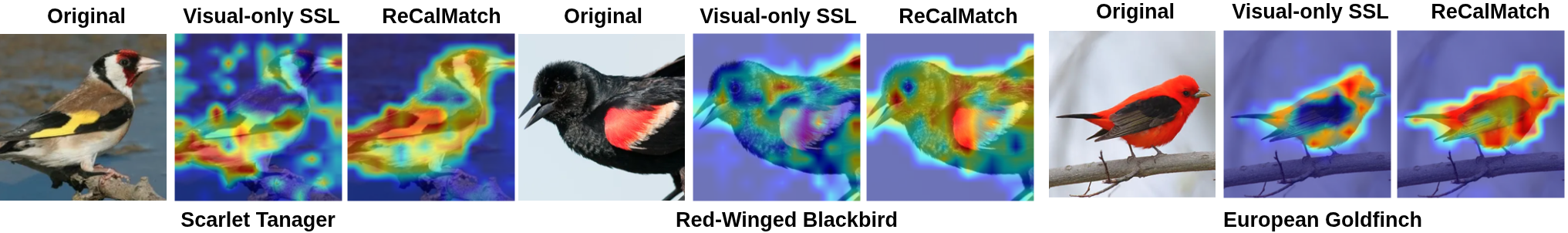}
    \vspace{-2em}
    \caption{Qualitative comparison on CUB-200-2011 under the 1\% labeled-data setting. Compared with a visual-only SSL baseline, ReCalMatch produces more object-aligned, class-discriminative activations and suppresses background-dominant responses.}
    \label{fig:teaser}
\end{figure}

The greatest difficulty is that \emph{high visual confidence is not a sufficient indicator of pseudo-label correctness in FGVC}. Because confusable categories share similar global appearance, a visual classifier can assign high confidence to the wrong class on an ambiguous sample. Mainstream SSL methods such as FixMatch~\cite{sohn2020fixmatch}, FlexMatch~\cite{A4}, and FreeMatch~\cite{wang2022freematch} decide whether to trust a pseudo-label based on signals derived purely from the visual prediction: maximum probability, fixed or adaptive thresholds~\cite{xu2021dash,A42}, and entropy statistics. When the classifier is overconfident on a visually ambiguous sample, these criteria still admit the incorrect pseudo-label, and the resulting gradient pushes the model further in the wrong direction. Consistent with this, realistic evaluations report that strong SSL methods provide limited or unstable gains on fine-grained benchmarks~\cite{su2021realistic,A46,A47,A48,A49,A50,A51,A52,A53,A54,A55,A56,A57,A58,A59,A60,A63}.

Vision-language priors offer an external source of evidence. CLIP aligns visual and textual representations in a shared space~\cite{radford2021learning}, which makes class names usable as lightweight semantic anchors. Prompt-based adaptations further specialize such priors for downstream recognition~\cite{zhou2022learning}. Recent semi-supervised FGVC methods, notably SemiVisBooster~\cite{zhang2025semivisbooster}, show that class-name semantics can improve pseudo-label learning. Yet these pipelines use semantics mainly as \emph{additional supervision}---aligning visual features toward text prototypes---rather than as a signal that decides \emph{whether a pseudo-label should be trusted at all}. Moreover, a single class-level text view is often too coarse for FGVC, where discriminative differences are distributed across several semantic aspects such as color, shape, and local parts.

We argue that pseudo-label reliability should be assessed \emph{jointly} based on three complementary signals: visual confidence, prediction uncertainty, and \textbf{visual--semantic agreement}---whether the unlabeled visual embedding is compatible with the semantic prototype of its predicted class. A pseudo-label that is visually confident but far from its class prototype in semantic space is precisely the kind of overconfident error that confidence-only criteria fail to catch. Building on this insight, we propose \textbf{ReCalMatch}, a reliability-calibrated semantic framework for semi-supervised FGVC. ReCalMatch uses a standard weak-to-strong SSL branch with a novel semantic calibration branch. It constructs class-conditioned multi-aspect semantic prototypes from class names and domain-specific aspects, computes a semantic agreement score for each pseudo-labeled sample, and uses this score together with confidence and entropy information to reweight the unsupervised consistency loss. A semantic consistency term aligns embeddings with calibrated prototypes, and a semantic margin regularizer reduces cross-class prototype redundancy so that the agreement signal stays discriminative. As illustrated in Figure~\ref{fig:teaser}, reliability calibration yields more object-aligned, class-discriminative responses and suppresses spurious background activations relative to a visual-only SSL baseline. Our contributions are:
\begin{itemize}
  \item We 
  show that visual-confidence-based acceptance is insufficient under small inter-class margins in semi-supervised FGVC under a \emph{pseudo-label reliability} perspective.
  \item We propose ReCalMatch, which uses multi-aspect semantic evidence to \emph{calibrate} pseudo-label reliability via a visual--semantic agreement score, rather than using semantics only as auxiliary supervision.
  \item We introduce a reliability-calibrated consistency loss, a semantic consistency term, and a semantic margin regularizer, and demonstrate consistent gains across four fine-grained benchmarks, with the largest improvements in low-label regimes.
\end{itemize}

\section{Related Work}

\subsection{Semi-Supervised Learning and Pseudo-Label Reliability}
SSL exploits unlabeled data through self-training and consistency regularization. Classical methods include Pseudo-Label~\cite{lee2013pseudo}, the $\Pi$-Model~\cite{laine2016temporal}, and Mean Teacher~\cite{tarvainen2017mean}, while later work couples pseudo-labeling with strong augmentation, as in MixMatch~\cite{berthelot2019mixmatch}, ReMixMatch~\cite{A31}, FixMatch~\cite{sohn2020fixmatch}, and UDA~\cite{xie2020unsupervised}. A central question is when to trust an unlabeled prediction. The dominant answer is to threshold or reweight pseudo-labels using statistics of the visual classifier itself: fixed confidence in FixMatch~\cite{sohn2020fixmatch}, curriculum thresholds in FlexMatch~\cite{A4}, dynamic thresholds in Dash~\cite{xu2021dash}, self-adaptive thresholds in FreeMatch~\cite{wang2022freematch}, distribution-aware reweighting in SoftMatch~\cite{A42}, and entropy minimization~\cite{A18}. AdaMatch~\cite{berthelot2021adamatch} further unifies thresholding across domains. These criteria substantially reduce label noise on generic benchmarks, but they share a common limitation: reliability is estimated entirely from the visual prediction. Confirmation bias has been studied as a direct consequence~\cite{A7}, and realistic evaluations confirm that gains can be unstable when inter-class margins are small~\cite{A2,su2021realistic}. ReCalMatch is motivated by this gap: instead of refining the threshold over the same visual signal, we introduce an \emph{external} reliability signal derived from semantic prototypes.

\subsection{Fine-Grained Visual Classification}
FGVC requires distinguishing categories that share global appearance but differ in subtle local cues~\cite{wei2021fine}. Early pipelines localize discriminative parts through bounding boxes or part annotations~\cite{zhang2014part}, while later CNN-based methods~\cite{fang2025low} learn attention or multi-region representations without explicit part labels~\cite{fu2017look,zheng2017learning,lin2015bilinear,gao2016compact,A28}. Token-level modeling with the Vision Transformer~\cite{dosovitskiy2020image} has further shifted FGVC toward global self-attention: TransFG selects discriminative tokens for fine-grained recognition~\cite{he2022transfg}, and subsequent work refines token interaction and multi-scale aggregation~\cite{wang2021feature,zhang2022free,hu2021rams,sun2022sim,xu2023fine}. Counterfactual~\cite{rao2021counterfactual} and cross-attention~\cite{zhu2022dual} variants further improve part-level discrimination. These methods drive supervised FGVC, but  are visual-only and do not provide a mechanism for assessing whether a pseudo-label predicted on an unlabeled image is trustworthy. Bringing SSL to FGVC is itself nontrivial: \cite{su2021realistic} reports that strong SSL methods can produce unstable gains on fine-grained benchmarks. Dedicated FGVC-SSL approaches such as self-training~\cite{nartey2019semi}, adversarial refinement~\cite{mugnai2022fine}, precision-enhanced pseudo-labeling~\cite{A32}, and soft label expansion/shrinkage~\cite{A33} attempt to stabilize this regime. ReCalMatch is complementary: rather than redesigning the visual pipeline, it injects an external semantic reliability signal into pseudo-label learning.
\begin{figure}[t]
    \centering\includegraphics[width=1\linewidth]{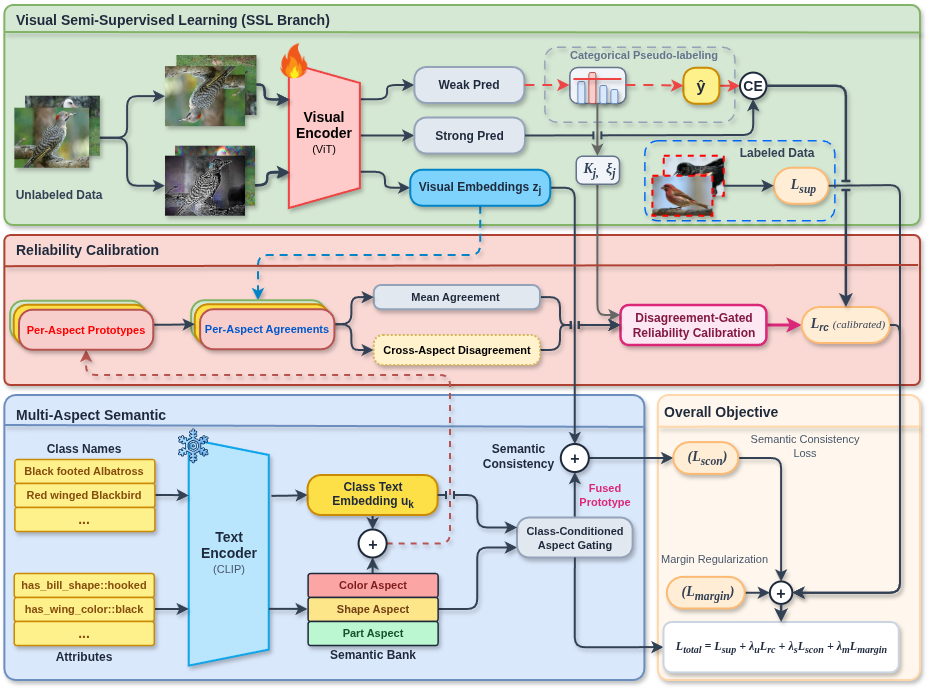}
    \vspace{-2em}
    \caption{Overview of \textbf{ReCalMatch}. The visual branch performs standard SSL, while the semantic branch constructs multi-aspect class prototypes from frozen CLIP text features. Cross-aspect agreement and disagreement are combined with classifier statistics to calibrate pseudo-label reliability and the unsupervised consistency loss.}
    \label{fig:pipeline}
\end{figure}
\subsection{Vision-Language Semantics for Recognition}
Large-scale vision-language pre-training; e.g.,  CLIP~\cite{radford2021learning},  provides class-name text embeddings that can serve as lightweight semantic anchors for visual recognition. Prompt-based adaptations specialize these anchors for downstream tasks~\cite{zhou2022learning,mao2023doubly}, and attribute-based representations~\cite{farhadi2009describing,lampert2013attribute} highlight that recognition benefits from semantic structure beyond a single class label. In semi-supervised FGVC, SemiVisBooster~\cite{zhang2025semivisbooster} uses class-name text embeddings to guide pseudo-label learning, and context-based semantic alignment has been explored for semi-supervised multi-label classification~\cite{fan2024context}. Part-level semantic-guided contrastive learning~\cite{linpart} and multi-grained cross-modal alignment for open-vocabulary segmentation~\cite{liu2024multi} further demonstrate that finer semantic granularity helps recognition. Existing semantic-guided SSL pipelines, however, share two limitations. First, semantics are used predominantly as \emph{auxiliary supervision}---visual features are pulled toward text prototypes---rather than as a criterion for deciding whether a pseudo-label should be trusted. Second, supervision is typically tied to a single class-level text view, which is too coarse for FGVC, where discriminative differences spread across multiple semantic aspects. ReCalMatch addresses both limitations by constructing multi-aspect semantic prototypes and using \emph{visual--semantic agreement} as an explicit calibration signal for pseudo-label reliability.

\section{ReCalMatch}

The overall architecture of ReCalMatch is shown in Figure~\ref{fig:pipeline}. The framework has two branches coupled by a reliability calibration block. The upper branch is a standard visual SSL pipeline that produces pseudo-labels and a confidence mask from weakly augmented unlabeled samples. The lower branch builds a multi-aspect semantic prior --- one prototype per aspect per class --- from class names and domain-specific aspect prompts. For each unlabeled sample, the visual embedding is compared against the per-aspect prototypes of its pseudo-label class to produce $V$ separately computed agreement signals; the mean and the cross-aspect disagreement of these signals are combined with the classifier confidence and with entropy information into a reliability weight that gates the unsupervised consistency loss. Two auxiliary terms --- a semantic consistency loss against a class-conditioned fused prototype and a prototype margin regularizer --- shape the prototype space so that the calibration signal remains discriminative. In this section we formalize each component.

\subsection{Baseline SSL and Pseudo-Label Reliability}
We consider a $K$-class semi-supervised setting. The labeled set and the unlabeled set are, respectively
\begin{equation}
\mathcal{D}_{\mathrm{L}}=\{(\mathbf{x}^{\mathrm{L}}_i,y_i)\}_{i=1}^{N_{\mathrm{L}}};  \space \mathcal{D}_{\mathrm{U}}=\{\mathbf{x}^{\mathrm{U}}_j\}_{j=1}^{N_{\mathrm{U}}},
\end{equation}
where $y_i\in\{1,\dots,K\}$. Let $f_{\theta}(\cdot)$ denote a visual network with parameters $\theta$, that outputs a logit vector and an embedding vector $\mathbf{z}\in\mathbb{R}^{d_z}$ for each image. For a mini-batch of labeled data $\mathcal{B}_{\mathrm{L}}$, the supervised term is
\begin{equation}
\mathcal{L}_{\mathrm{sup}}=\frac{1}{|\mathcal{B}_{\mathrm{L}}|}\sum_{(\mathbf{x},y)\in\mathcal{B}_{\mathrm{L}}}\mathrm{CE}\big(f_{\theta}(\omega(\mathbf{x})),y\big),
\end{equation}
where $\omega(\cdot)$ denotes weak augmentation, and CE is the cross entropy. For each unlabeled sample $\mathbf{x}^{\mathrm{U}}_j$, we generate a weak view $\omega(\mathbf{x}^{\mathrm{U}}_j)$ and a strong view $\mathcal{A}(\mathbf{x}^{\mathrm{U}}_j)$. The weak-view class distribution $\mathbf{p}_j$ with pseudo-label $\hat{y}_j$ is
\begin{equation}
\mathbf{p}_j=\mathrm{softmax}\big(f_{\theta}(\omega(\mathbf{x}^{\mathrm{U}}_j))\big),\space
\hat{y}_j=\arg\max_{c\in\{1,\dots,K\}}p_{j,c}.
\end{equation}

Let $\mu_j\in\{0,1\}$ be the confidence mask produced by the underlying SSL backbone (e.g., FixMatch~\cite{sohn2020fixmatch} or FreeMatch~\cite{wang2022freematch}), and let $\kappa_j=\max_c p_{j,c}$ and $\xi_j=-\frac{1}{\log K}\sum_{c=1}^{K}p_{j,c}\log p_{j,c}$ denote,  respectively, the classifier confidence and normalized entropy. Both the mask and these statistics are derived entirely from the visual prediction. In FGVC, confusable categories share global appearance, and the classifier can be confidently wrong on visually ambiguous samples; the mask alone cannot distinguish a high-confidence correct prediction from an overconfident pseudo-label error. ReCalMatch therefore introduces an \emph{external} reliability signal --- one that does not depend on the visual classifier's own statistics --- by querying multiple independent semantic aspects of the predicted class.

\subsection{Multi-Aspect Semantic Prior Construction}

We construct the semantic prior offline from two sources: a class-name text embedding and a multi-aspect semantic bank. Crucially, two distinct prototype constructions are derived from these sources and serve disjoint roles in the rest of the framework.

Let $\psi(\cdot)$ denote a frozen text encoder. For class $k$ with name string $\tau_k$, the class-name embedding is
\begin{equation}
\mathbf{t}^{\mathrm{cls}}_k=\psi(\tau_k)\in\mathbb{R}^{d_t},\quad k=1,\dots,K.
\end{equation}
We maintain $V$ semantic aspects, each describing a complementary facet of the category (e.g., color, shape, and local parts for biological FGVC). Supervision at different semantic granularities provides complementary cues~\cite{liu2024multi}. Each aspect contains $P$ prompts; the embedding of $\rho_{v,p}$; i.e., the $p$-th prompt of aspect $v$ is
\begin{equation}
\mathbf{t}^{\mathrm{asp}}_{v,p}=\psi(\rho_{v,p})\in\mathbb{R}^{d_t},\quad v=1,\dots,V,\;p=1,\dots,P.
\end{equation}
A learnable projector $\phi:\mathbb{R}^{d_t}\rightarrow\mathbb{R}^{d_z}$ maps text embeddings into the visual feature space:
\begin{equation}
\mathbf{u}_k=\phi(\mathbf{t}^{\mathrm{cls}}_k),\qquad \mathbf{q}_{v,p}=\phi(\mathbf{t}^{\mathrm{asp}}_{v,p}),
\end{equation}
and each aspect is summarized by prompt pooling:
\begin{equation}
\mathbf{g}_v=\frac{1}{P}\sum_{p=1}^{P}\mathbf{q}_{v,p}\in\mathbb{R}^{d_z}.
\end{equation}

\textbf{Per-aspect class prototypes.} For the reliability calibration mechanism in Sec.~\ref{sec:agreement}, we treat each aspect as a separately constructed piece of semantic evidence. We therefore define a per-aspect class prototype as the simple additive combination of the class-name embedding and the aspect prototype:
\begin{equation}
\mathbf{m}_k^{(v)}=\mathbf{u}_k+\mathbf{g}_v\in\mathbb{R}^{d_z},\quad v=1,\dots,V.
\label{eq:peraspect}
\end{equation}
No cross-aspect fusion is applied at this stage. Each aspect contributes its own view of class $k$ and produces its own agreement signal.

\textbf{Fused class prototype.} For the auxiliary semantic consistency loss (Sec.~\ref{sec:auxloss}), a single \emph{consensus} target per class is needed. We obtain it through class-conditioned aspect gating:
\begin{equation}
\eta_{k,v}=\frac{\exp\big(\mathrm{sim}(\mathbf{u}_k,\mathbf{g}_v)/\tau_{g}\big)}{\sum_{r=1}^{V}\exp\big(\mathrm{sim}(\mathbf{u}_k,\mathbf{g}_r)/\tau_{g}\big)},
\end{equation}
\begin{equation}
\mathbf{s}_k=\mathbf{u}_k+\sum_{v=1}^{V}\eta_{k,v}\,\mathbf{g}_v\in\mathbb{R}^{d_z},
\label{eq:fused}
\end{equation}
where $\mathrm{sim}(\cdot,\cdot)$ is cosine similarity and $\tau_{g}>0$ is a gate temperature. The fused prototype $\mathbf{s}_k$ is used \emph{only} by the auxiliary losses; its class-conditioned gate is intentionally image-independent to provide a stable class-level consensus target, while image-specific adaptivity is supplied by the per-sample agreements $a_j^{(v)}$, their mean $\bar{a}_j$, and disagreement $d_j$ used for reliability calibration.

\subsection{Semantic Bank Instantiation and Cross-Dataset Protocol}
\label{sec:semantic}
The semantic bank is constructed \emph{offline} at the class level and stored as a fixed tensor of shape $[V,P,d_t]$. This precomputation ensures deterministic supervision across runs, removes repeated text encoding from the optimization loop, and provides a modular interface that plugs into different SSL backbones without modification.

For \textbf{CUB-200-2011}~\cite{A13}, the bank is built directly from official attribute and part annotations: deterministic keyword rules map attribute entries to color and shape aspects, while part annotations populate the part aspect. Parsed entries are wrapped in fixed natural-language templates, encoded by $\psi$, projected by $\phi$, and normalized to a common prompt count per aspect by truncation or repetition, yielding $V=3$. For \textbf{Stanford Dogs}~\cite{standforddogs}, we use class-name-driven template expansion with canine-specific color, shape, and part terminology. For \textbf{NABirds}~\cite{A12}, we reuse the bird-oriented templates and part vocabulary established in the CUB protocol. For \textbf{iNaturalist18}~\cite{A11}, we combine class names with general color and shape templates and a generic part aspect to preserve the unified three-aspect interface across domains. Color, shape, and part are biological-FGVC instantiations of the bank interface; for a new domain, the same fixed $[V,P,d_t]$ interface is retained while only the class-name-driven aspect prompts are replaced by domain-relevant descriptions (e.g., manufacturer, body type, and component layout for vehicles), requiring no per-image semantic annotation.

\subsection{Per-Aspect Agreement and Cross-Aspect Disagreement}
\label{sec:agreement}
The central calibration mechanism of ReCalMatch is the use of \emph{disagreement} among independent semantic aspects as a reliability signal for pseudo-labels. We first compute one agreement score per aspect, then aggregate the scores into a mean and a cross-aspect disagreement. For an unlabeled sample $\mathbf{x}^{\mathrm{U}}_j$ with embedding $\mathbf{z}_j$ and pseudo-label $\hat{y}_j$, the agreement score of aspect $v$ is
\begin{equation}
a_j^{(v)}=\frac{\exp\big(\mathrm{sim}(\mathbf{z}_j,\mathbf{m}_{\hat{y}_j}^{(v)})/\tau_{s}\big)}{\sum_{c=1}^{K}\exp\big(\mathrm{sim}(\mathbf{z}_j,\mathbf{m}_{c}^{(v)})/\tau_{s}\big)},
\label{eq:peraspect_agreement}
\end{equation}
where $\tau_{s}>0$ is the agreement temperature. Each $a_j^{(v)}$ answers a localized question: \emph{does aspect $v$ alone support the pseudo-label $\hat{y}_j$?} Because the $V$ scores are computed separately against the per-aspect prototypes from Eq.~\ref{eq:peraspect}, they provide parallel semantic tests of the same pseudo-label. Note that this construction does not assume that the underlying aspects are statistically independent. We then summarize the $V$ scores by their first and second moments. The mean agreement, $\bar{a}_j$, captures the average semantic support for $\hat{y}_j$, while the cross-aspect disagreement, $d_j$, 
measures the extent to which independent aspects fail to concur:
\begin{equation}
\bar{a}_j=\frac{1}{V}\sum_{v=1}^{V}a_j^{(v)}, \space\text{  } d_j=\frac{1}{V}\sum_{v=1}^{V}\big(a_j^{(v)}-\bar{a}_j\big)^2.
\label{eq:disagreement}
\end{equation} 
A sample on which all aspects assign high agreement to $\hat{y}_j$ has high $\bar{a}_j$ and low $d_j$; 
a sample on which one aspect agrees strongly while the others provide weak support for  $\hat{y}_j$ has moderate $\bar{a}_j$ but high $d_j$.
The latter pattern is precisely the failure mode targeted by ReCalMatch: a pseudo-label that the visual classifier endorses with high confidence but on which independent semantic evidence is internally inconsistent.

The \emph{disagreement-gated} reliability calibration weight combines visual statistics with these two semantic summaries:
\begin{equation}
r_j=\kappa_j^{\alpha}\,(1-\xi_j)^{\beta}\,\bar{a}_j^{\gamma}\,\exp\!\big(-\zeta\,d_j\big),
\label{eq:reliability}
\end{equation}
where $\alpha,\beta,\gamma,\zeta\ge 0$ control the contribution of each factor. The first two factors are visual-only and reproduce conventional confidence-based reliability criteria. The third factor injects the mean semantic agreement. The fourth factor, $\exp(-\zeta\,d_j)$, is the distinguishing component of ReCalMatch: it exponentially suppresses pseudo-labels on which independent aspects disagree, even when $\kappa_j$, $\xi_j$, and $\bar{a}_j$  individually accept them. Because the factor is bounded above by $1$ and approaches $0$ as $d_j$ grows, it acts as a smooth, differentiable gate on cross-aspect inconsistency.

\subsection{Reliability-Calibrated Consistency Learning}
Instead of applying the unsupervised consistency loss uniformly to every pseudo-label admitted by $\mu_j$, ReCalMatch reweights it by the calibration weight $r_j$:
\begin{equation}
\mathcal{L}_{\mathrm{rc}}=\frac{1}{|\mathcal{B}_{\mathrm{U}}|}\sum_{j\in\mathcal{B}_{\mathrm{U}}}\mu_j\,r_j\,\mathrm{CE}\big(f_{\theta}(\mathcal{A}(\mathbf{x}^{\mathrm{U}}_j)),\hat{y}_j\big).
\label{eq:Lrc}
\end{equation}
When all four factors of $r_j$ are near unity --- the pseudo-label is confident, certain, semantically well-supported on average, and consistently supported across aspects --- $\mathcal{L}_{\mathrm{rc}}$ recovers the standard unsupervised consistency loss. When any factor deteriorates, the corresponding sample's gradient contribution is smoothly suppressed. The disagreement gate $\exp(-\zeta\,d_j)$ catches the specific failure mode that the other three factors miss: high visual confidence with internally inconsistent semantic support.

\subsection{Auxiliary Semantic Consistency and Margin Regularization}
\label{sec:auxloss}
For the agreement scores $a_j^{(v)}$ to be informative, two conditions must hold throughout training: visual embeddings must be drawn toward the semantic prototypes of their (predicted) class, and the prototypes themselves must be sufficiently separated. Two auxiliary losses enforce these conditions.

\textbf{Semantic consistency.} Using the fused class prototype $\mathbf{s}_k$ from Eq.~\ref{eq:fused} as the consensus target, the per-sample semantic consistency loss is
\begin{equation}
\ell_{i}^{\mathrm{scon}}=-\log\frac{\exp\big(\mathrm{sim}(\mathbf{z}_i,\mathbf{s}_{\bar{y}_i})/\tau_{c}\big)}{\sum_{c=1}^{K}\exp\big(\mathrm{sim}(\mathbf{z}_i,\mathbf{s}_c)/\tau_{c}\big)},
\end{equation}
where $\bar{y}_i=y_i$ for labeled samples, $\bar{y}_i=\hat{y}_i$ for unlabeled samples, and $\tau_{c}>0$ is the consistency temperature. Labeled samples carry unit weight; unlabeled samples are reweighted by $\mu_j r_j$, so that pseudo-labels that fail the reliability calibration do not pull the embedding toward an incorrect prototype:
\begin{equation}
\mathcal{L}_{\mathrm{scon}}=\frac{1}{|\mathcal{B}_{\mathrm{L}}|+|\mathcal{B}_{\mathrm{U}}|}\!\left(\sum_{i\in\mathcal{B}_{\mathrm{L}}}\ell_{i}^{\mathrm{scon}}+\sum_{j\in\mathcal{B}_{\mathrm{U}}}\mu_j\,r_j\,\ell_{j}^{\mathrm{scon}}\right).
\end{equation}

\textbf{Margin regularization.}  If two fused class prototypes are highly correlated, the semantic space provides weak inter-class separation. 
Since the fused and per-aspect prototypes share the same class embedding and aspect representations, weak prototype separation can also reduce the discriminability of the per-aspect agreement scores $a_j^{(v)}$ in Eq.~\ref{eq:peraspect_agreement}, consequently weakening the disagreement signal $d_j$ in Eq.~\ref{eq:disagreement}. To prevent this, we normalize each fused prototype $\tilde{\mathbf{s}}_k=\mathbf{s}_k/\|\mathbf{s}_k\|_2$, form the correlation matrix $\mathbf{R}=\tilde{\mathbf{S}}\tilde{\mathbf{S}}^{\top}\in\mathbb{R}^{K\times K}$ with $\tilde{\mathbf{S}}=[\tilde{\mathbf{s}}_1,\dots,\tilde{\mathbf{s}}_K]^{\top}$, and penalize off-diagonal correlations beyond a margin $\delta_{s}\ge 0$:
\begin{equation}
\mathcal{L}_{\mathrm{margin}}=\frac{1}{K(K-1)}\sum_{k=1}^{K}\sum_{\substack{l=1\\l\neq k}}^{K}\max\!\big(|R_{k,l}|-\delta_{s},\,0\big).
\label{eq:margin}
\end{equation}
This penalty acts directly on the fused prototype space and, through the shared class and aspect representations in Eq.~\ref{eq:peraspect}, indirectly improves the separability of the per-aspect agreement scores defined in Eq.~\ref{eq:peraspect_agreement}.
Note that $\delta_s=0$ does not disable the regularizer: it reduces Eq.~\ref{eq:margin} to penalizing the absolute off-diagonal prototype correlations, while larger $\delta_s$ progressively relaxes this decorrelation constraint.

\subsection{Overall Objective and Training Protocol}
The full ReCalMatch objective combines the supervised loss, the reliability-calibrated consistency loss, and the two auxiliary terms:
\begin{equation}
\mathcal{L}_{\mathrm{total}}=\mathcal{L}_{\mathrm{sup}}+\lambda_{u}\,\mathcal{L}_{\mathrm{rc}}+\lambda_{s}\,\mathcal{L}_{\mathrm{scon}}+\lambda_{m}\,\mathcal{L}_{\mathrm{margin}},
\label{eq:total}
\end{equation}
where $\lambda_{u},\lambda_{s},\lambda_{m}$ are scalar trade-off coefficients. ReCalMatch is implemented as a lightweight calibration branch on top of a standard SSL pipeline: the baseline pseudo-labeling and consistency machinery (e.g., FixMatch~\cite{sohn2020fixmatch} or FreeMatch~\cite{wang2022freematch}) is preserved, while the calibration branch contributes the per-aspect agreement scores $a_j^{(v)}$, the disagreement $d_j$, the reliability weight $r_j$, the auxiliary semantic consistency, and the prototype margin regularizer. Because the semantic bank is constructed offline and the calibration block operates on a fixed $K\times d_z$ prototype matrix per aspect, the per-iteration cost added by the calibration branch is dominated by inexpensive cosine similarity computations.

\section{Experiments and Results}

\subsection{Implementation Details}
\label{sec:implementation}
All models are implemented in PyTorch on the USB benchmark~\cite{usb2022} and trained on a single GPU. For ReCalMatch, we use ViT-B/16~\cite{dosovitskiy2020image} as the visual backbone with input resolution $224\times224$ and standard normalization, while competing methods follow their default settings unless otherwise specified. We do not adopt the CLIP visual encoder as the default backbone, since its stronger vision-language pretraining would confound the contribution of the proposed calibration mechanism in fair SSL comparisons. Weak augmentations include random crop and horizontal flip, and strong augmentations follow RandAugment, consistent with common SSL practice.

Training follows the FreeMatch protocol with a temperature of 0.5 and an EMA momentum of 0.999 for updating the probability model. We use AdamW with a learning rate of $5\times10^{-4}$, cosine decay, a weight decay of $5\times10^{-4}$, and a warm-up phase of 5{,}120 iterations. During early training, the warm-up together with the reliability factor $\mu_jr_j$ limits the influence of uncertain pseudo-labels on $\mathcal{L}_{\mathrm{scon}}$; moreover, the text encoder $\psi$ remains frozen and only the projector $\phi$ is adapted. Unless otherwise specified, hyperparameters are shared across datasets and label ratios. The batch size is 8 for training and 16 for evaluation, models are trained for 200 epochs, and evaluation uses the EMA model. Unless otherwise stated, the random seed is 0.

For the reliability calibration weight $r_j$ in Eq.~\ref{eq:reliability}, we use confidence and entropy exponents $\alpha=\beta=1.0$, mean-agreement exponent $\gamma=1.0$, and disagreement scale $\zeta=2.0$. Temperatures are set to $\tau_g=\tau_s=\tau_c=0.07$, and the prototype margin in $\mathcal{L}_{\mathrm{margin}}$ is $\delta_s=0$. Trade-off coefficients in Eq.~\ref{eq:total} are $\lambda_u=1.0$, $\lambda_s=1.0$, and $\lambda_m=0.05$. The semantic bank uses $V=3$ aspects (color, shape, part), with the per-aspect prompt count $P$ normalized per dataset as described in Sec.~\ref{sec:semantic}. For large-$K$ datasets such as iNaturalist18, the $K\times K$ prototype-correlation term in $\mathcal{L}_{\mathrm{margin}}$ is evaluated on a sampled class subset per step rather than materializing the full matrix.

\subsection{Datasets}
\label{sec:datasets}
We evaluate ReCalMatch on four fine-grained benchmarks: iNaturalist18~\cite{A11}, NABirds~\cite{A12}, CUB-200-2011~\cite{A13}, and Stanford Dogs~\cite{standforddogs}. These datasets cover subtle inter-class variations (a core property of FGVC) while presenting diverse degrees of class imbalance, which is crucial for validating robustness in realistic settings. Following prior work, we report results under 1\%, 10\%, and 25\% labeled-data settings, where labeled subsets are sampled with a fixed seed and all remaining images are used as unlabeled data.

\textbf{CUB-200-2011 (CUB)}~\cite{A13} is a standard FGVC benchmark consisting of 200 bird species with 11{,}788 images. It exhibits \emph{moderate} class imbalance and remains challenging due to subtle inter-class differences and large intra-class variation. \textbf{Stanford Dogs}~\cite{standforddogs} contains 120 dog breeds with 20{,}580 images collected from ImageNet. Each class typically contains around 150--200 images, yielding a \emph{moderately} imbalanced distribution compared with long-tailed benchmarks. \textbf{NABirds}~\cite{A12} includes 555 North American bird species with 48{,}562 images. Its distribution is \emph{severely} imbalanced, where some species have hundreds of samples while many tail species have fewer than 50 images, making it a representative long-tailed FGVC benchmark. \textbf{iNaturalist18}~\cite{A11} is a large-scale real-world benchmark with 8{,}142 categories and approximately 437{,}000 images. It exhibits a highly long-tailed distribution, with substantial frequency gaps between head and tail classes, and is used to evaluate robustness under extreme class imbalance.

\subsection{Comparison with the State of the Art}

\begin{table*}
\centering
\small
\scriptsize
\caption{Performance of SSL methods on fine-grained datasets. Top-1 accuracy (\%) under different labeled-data settings. The \textbf{bold} and \underline{underlined} figures represent the best and second-best results, respectively.}
\label{tab:main_results}
\resizebox{\linewidth}{!}{
\begin{tabular}{l|c|ccc|ccc|ccc|ccc}
\toprule
\multirow{2}{*}{\textbf{Method}} &
\multirow{2}{*}{\textbf{Param.(M)}} &
\multicolumn{3}{c|}{\textbf{CUB-200-2011}} &
\multicolumn{3}{c|}{\textbf{Stanford Dogs}} &
\multicolumn{3}{c|}{\textbf{iNaturalist18}} &
\multicolumn{3}{c}{\textbf{NABirds}} \\
 & & 1\% & 10\% & 25\% & 1\% & 10\% & 25\% & 1\% & 10\% & 25\% & 1\% & 10\% & 25\% \\
\midrule
Pseudo-Label\cite{lee2013pseudo}              & 25  & 32.1 & 58.5 & 68.3 & 29.4 & 56.2 & 66.5 & 34.2 & 55.0 & 65.9 & 30.7 & 55.3 & 65.2 \\
FreeMatch\cite{wang2022freematch}             & 26  & 35.6 & 63.4 & 72.1 & 33.2 & 60.5 & 70.0 & 36.9 & 59.4 & 70.2 & 32.8 & 58.7 & 68.9 \\
FixMatch\cite{sohn2020fixmatch}               & 25  & 40.9 & 65.7 & 74.2 & 36.5 & 62.8 & 72.1 & 39.3 & 61.1 & 72.8 & 35.8 & 61.9 & 71.4 \\
FlexMatch\cite{A4}                            & 85  & 44.5 & 71.3 & 80.5 & 42.2 & 69.1 & 78.3 & 45.2 & 67.8 & 78.4 & 41.6 & 67.8 & 77.2 \\
SelfMatch\cite{A5}                            & 80  & 46.3 & 73.1 & 81.3 & 44.8 & 71.5 & 80.2 & 47.1 & 69.5 & 79.9 & 43.7 & 69.4 & 78.8 \\
Semi-ViT-Huge\cite{A20}                       & 631 & 52.0 & 76.5 & 82.7 & \underline{50.1} & 72.6 & \underline{82.5} & 50.9 & 72.2 & 82.9 & \underline{48.2} & \underline{73.1} & \underline{81.9} \\
MoCo + SimMatch\cite{zheng2022simmatch}                     & 127 & 52.0 & 76.3 & 82.5 & 48.5 & \underline{74.6} & 81.0 & 46.2 & 68.1 & 82.3 & 45.4 & 70.8 & 81.2 \\
MoCo + SoC\cite{A33}                          & 89  & 50.7 & \underline{77.4} & \underline{83.1} & 48.2 & 73.3 & 80.7 & \underline{52.4} & \underline{74.3} & \textbf{83.4} & 47.6 & 72.5 & 81.2 \\ 
PEPL\cite{A32}                                & 25  & 51.3 & 76.1 & \textbf{83.2} & 46.8 & 74.1 & 82.3 & 51.6 & 73.0 & 82.1 & 46.5 & 71.2 & 80.3 \\
SemiVisBooster\cite{zhang2025semivisbooster}  & 127 & \underline{53.2} & 74.2 & 81.7 & 47.1 & 72.2 & 81.9 & 47.3 & 69.6 & 82.8 & 45.8 & 71.0 & 81.4 \\
SemiViM\cite{he2025semi}                      & 27  & 48.2 & 75.7 & 82.2 & 46.5 & 72.8 & 82.1 & 47.7 & 70.2 & 79.1 & 46.6 & 72.1 & 80.1 \\
\midrule
\textbf{ReCalMatch (Ours)}                    & 128 & \textbf{54.2} & \textbf{78.2} & 82.3
                                                    & \textbf{52.2} & \textbf{76.1} & \textbf{82.7}
                                                    & \textbf{54.2} & \textbf{74.7} & \underline{83.2}
                                                    & \textbf{49.9} & \textbf{74.0} & \textbf{82.1} \\
\bottomrule
\end{tabular}
}
    \vspace{-1em}
\end{table*}

Table~\ref{tab:main_results} reports the results of SSL methods under 1\%, 10\%, and 25\% labeled-data settings on the four fine-grained benchmarks. Overall, ReCalMatch achieves the strongest performance in the most challenging low-label regimes and remains highly competitive as the labeled ratio increases. The gains are most pronounced at the 1\% and 10\% labeled-data settings, where pseudo-label ambiguity is most severe and overconfident pseudo-label errors are most likely to be reinforced through consistency learning. This trend directly supports the central motivation of the framework: when visual evidence alone is insufficient to verify a pseudo-label, an external reliability signal derived from cross-aspect disagreement filters precisely the cases that confidence-based criteria admit by mistake. Compared with the strongest text-guided baseline, SemiVisBooster~\cite{zhang2025semivisbooster}, which uses a single class-name semantic target, ReCalMatch consistently improves accuracy across all four benchmarks, indicating that decomposing semantic supervision into independent aspects and exploiting their disagreement is more effective than tying semantics to a single supervision target. At the 25\% labeled-data setting, the gap between methods naturally narrows as stronger supervision reduces pseudo-label noise, yet ReCalMatch remains robust and competitive across all datasets. These results show that the proposed calibration mechanism is particularly beneficial in low-label semi-supervised FGVC, while preserving strong accuracy when more labeled data are available.

\subsection{Visualization Results}

\begin{figure*}[t]
    \centering
    \includegraphics[width=1\linewidth]{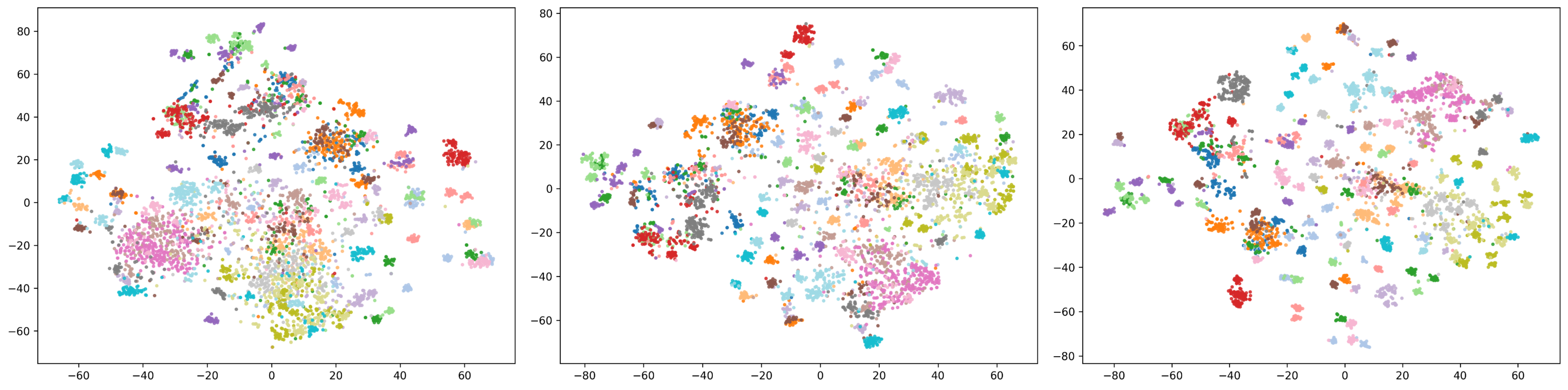}
     \vspace{-2em}
    \caption{t-SNE visualization of feature embeddings produced by different methods on CUB-200-2011. Left to right: visual-only SSL baseline (FreeMatch), SemiVisBooster with class-level text guidance, and ReCalMatch with reliability-calibrated semantic guidance.}
    \label{fig:tsnemap}
   
\end{figure*}

\begin{figure*}[t]
    \centering
    \includegraphics[width=1\linewidth]{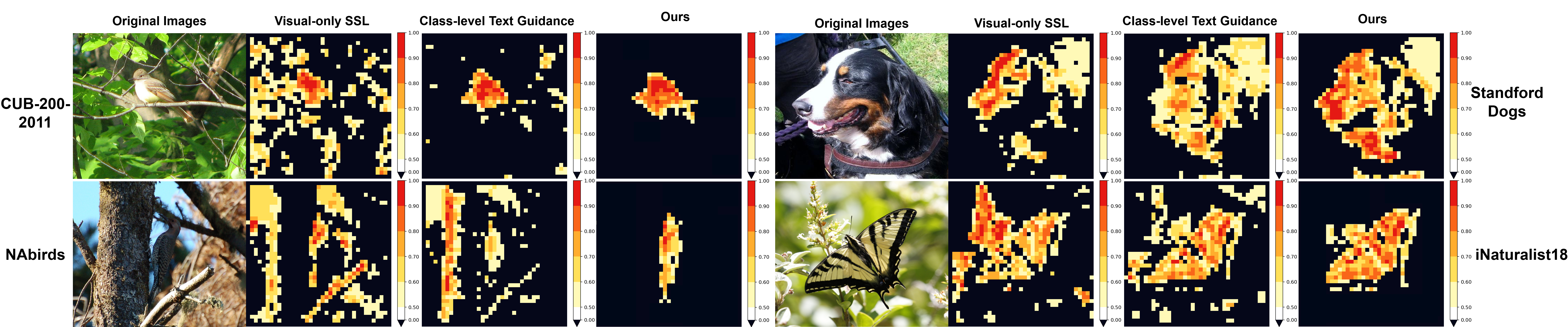}
    \vspace{-1.8em}
    \caption{Qualitative comparison of discriminative region localization across models with different semantic supervision settings. Left to right: input image, visual-only SSL baseline (FreeMatch), SemiVisBooster with class-level text guidance, and ReCalMatch with reliability-calibrated semantic guidance.}
    \label{fig:attnmap}
\end{figure*}

To analyze the learned representation space, we visualize feature embeddings with t-SNE. As shown in Figure~\ref{fig:tsnemap}, the visual-only baseline yields relatively entangled feature distributions with substantial overlap across categories. Incorporating class-level text supervision improves the global organization of the embedding space, yet several clusters remain diffuse and partially mixed. By comparison, ReCalMatch produces a more compact and better-separated feature geometry, with clearer local cluster formation and reduced cross-class interference. This qualitative evidence supports that suppressing pseudo-labels with high cross-aspect disagreement yields representations whose class structure is more consistent with the semantic prior than representations learned under a single semantic target.

We further compare activation maps to analyze how reliability calibration affects spatial attention. As shown in Figure~\ref{fig:attnmap}, the visual-only baseline tends to produce scattered responses and is easily distracted by background content. SemiVisBooster partially alleviates this issue through class-level text guidance, but its activations remain relatively diffuse. ReCalMatch generates more compact and object-aligned responses on discriminative fine-grained regions, indicating that calibrating consistency learning by cross-aspect disagreement encourages the model to attend to image regions on which independent semantic aspects consensually support the predicted class, while suppressing background-dominant responses on which the aspects disagree.

\section{Ablation Studies}
\label{sec:ablation}

\subsection{Component and Calibration Analysis}
We ablate ReCalMatch on CUB-200-2011 under the 1\%, 10\%, and 25\% labeled-data settings; all other settings follow Section~\ref{sec:implementation}.

\begin{wraptable}{r}{0.48\linewidth}\centering\vspace{-0.8em}\small\setlength{\tabcolsep}{4pt}\renewcommand{\arraystretch}{0.92}\caption{Component ablation (Top-1 accuracy -- \%) on CUB-200-2011 under
different labeled-data settings.}\label{tab:main_ablation}\begin{tabular}{l|ccc}\toprule\textbf{Configuration} & \textbf{1\%} & \textbf{10\%} & \textbf{25\%} \\\midrule FreeMatch & 35.6 & 63.4 & 72.1 \\ + $\mathcal{L}_{\mathrm{scon}}$ (class-name) & 50.9 & 74.8 & 80.5 \\ + multi-aspect prototypes & 52.3 & 77.3 & 81.2 \\ + $\mathcal{L}_{\mathrm{margin}}$ & 53.0 & 77.6 & 81.6 \\ + reliability calibration & 53.5 & 77.9 & 81.9 \\ + disagreement gating (full) & \textbf{54.2} & \textbf{78.2} & \textbf{82.3} \\\bottomrule\end{tabular}\vspace{-0.8em}\end{wraptable}
Table~\ref{tab:main_ablation} progressively builds ReCalMatch from the FreeMatch baseline. Adding the semantic consistency loss $\mathcal{L}_{\mathrm{scon}}$ with class-name prototypes already yields a large gain, and replacing them with multi-aspect prototypes improves further by incorporating complementary color, shape, and part cues. The margin regularizer $\mathcal{L}_{\mathrm{margin}}$ sharpens the prototype space, and reliability calibration based on confidence, entropy, and mean semantic agreement adds a further increment. Finally, the cross-aspect disagreement gating $\exp(-\zeta d_j)$ in Eq.~\ref{eq:reliability} --- the distinguishing component of ReCalMatch --- contributes the last improvement, with the largest absolute gain at 1\% labels, where overconfident pseudo-label errors most strongly dominate the consistency loss. The framework does not depend on a specific text encoder: replacing CLIP with FastText, BERT, or T5 yields a 53.6\%, 52.9\%, and 52.8\% accuracy, respectively,  at 1\% labels, all within 1.4\% of the default 54.2\% accuracy, confirming that the gain comes from external semantic calibration rather than a particular encoder. We also verify the consensus-target construction: uniform averaging and a global learnable fusion reach only 52.3\% and 53.0\% accuracy, respectively, at 1\% labels, versus 54.2\% accuracy for the class-conditioned aspect gating of Eq.~\ref{eq:fused}. This gating affects only the fused prototype $\mathbf{s}_k$ used by $\mathcal{L}_{\mathrm{scon}}$; the per-aspect prototypes $\mathbf{m}_k^{(v)}$ that drive the calibration mechanism remain ungated.

Beyond accuracy, the disagreement gate directly improves pseudo-label quality on CUB-200-2011 at 1\% labels: accepted-pseudo-label precision increases from $71.3\%$ to $78.6\%$, while retaining $92\%$ of correct pseudo-labels, confirming that the gate preferentially removes overconfident errors rather than simply reducing pseudo-label coverage.

\begin{figure}[t]
\centering
\begin{minipage}[h]{0.48\linewidth}
\centering
\includegraphics[width=1\linewidth]{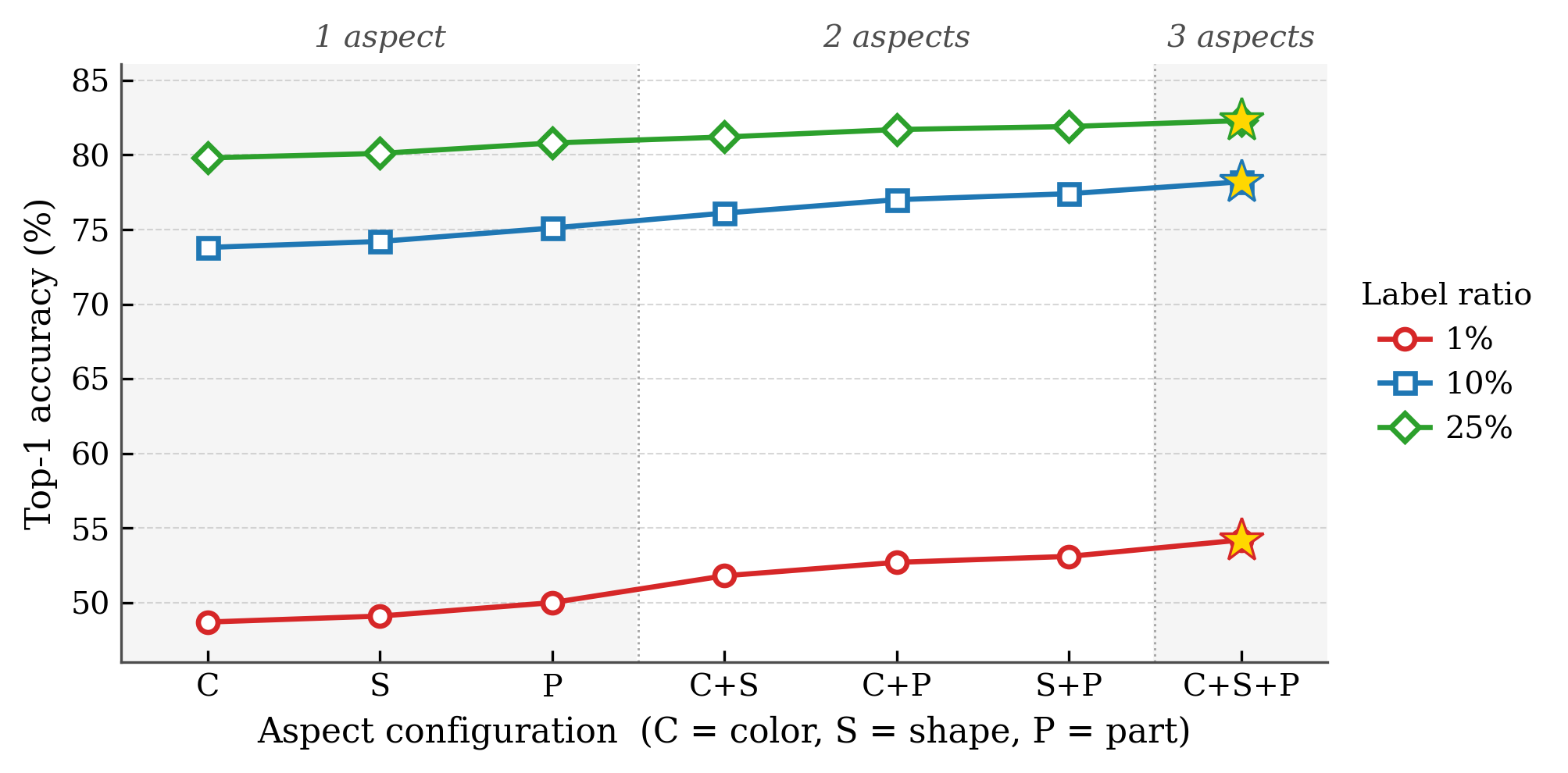}
\vspace{-2em}
\caption{Effect of semantic aspect configuration on CUB-200-2011.}
\label{fig:aspect_config}
\end{minipage}\hfill
\begin{minipage}[h]{0.48\linewidth}
\centering
\includegraphics[width=1\linewidth]{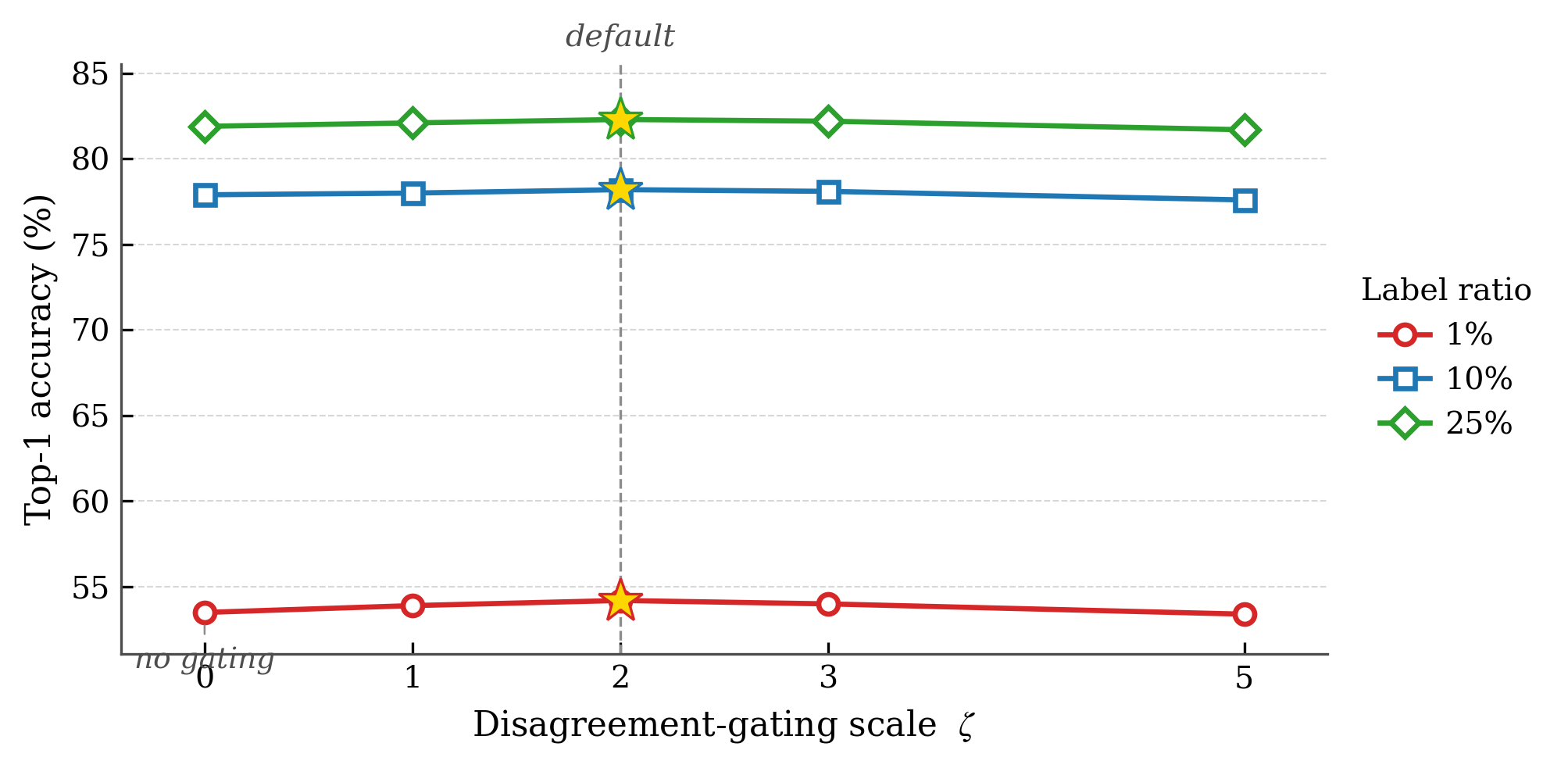}
\vspace{-2em}
\caption{Sensitivity to the disagreement-gating scale $\zeta$.}
\label{fig:zeta_sensitivity}
\end{minipage}
\end{figure}
\vspace{-1em}
\begin{table}[t]
\centering
\small
\caption{Compact sensitivity and stability analysis on CUB-200-2011 -- Top-1 accuracy (\%). Multi-seed rows report mean\,$\pm$\,std over three runs.}
\label{tab:sensitivity}
\begin{tabular}{l|l|ccc}
\toprule
\textbf{Group} & \textbf{Setting} & \textbf{1\%} & \textbf{10\%} & \textbf{25\%} \\
\midrule
\multirow{5}{*}{$(\lambda_s,\lambda_m)$}
 & (0.1, 0.05) & 51.9 & 76.6 & 81.4 \\
 & (0.5, 0.05) & 53.4 & 77.8 & 82.0 \\
 & (1.0, 0.05) & \textbf{54.2} & \textbf{78.2} & \textbf{82.3} \\
 & (2.0, 0.05) & 53.7 & 77.9 & 82.1 \\
 & (1.0, 0.10) & 54.0 & 78.1 & 82.2 \\
\midrule
\multirow{6}{*}{$(\alpha,\beta)$}
 & (0.0, 0.0) & 51.5 & 76.8 & 81.1 \\
 & (1.0, 0.0) & 52.4 & 77.4 & 81.7 \\
 & (0.0, 1.0) & 52.8 & 77.3 & 81.6 \\
 & (0.5, 0.5) & 53.6 & 77.9 & 82.1 \\
 & (1.0, 1.0) & \textbf{54.2} & \textbf{78.2} & \textbf{82.3} \\
 & (2.0, 2.0) & 53.5 & 77.7 & 81.9 \\
\midrule
\multirow{3}{*}{Seeds}
 & FreeMatch              & 35.6\,$\pm$\,0.4 & 63.4\,$\pm$\,0.4 & 72.1\,$\pm$\,0.3 \\
 & w/o disagreement       & 53.5\,$\pm$\,0.4 & 77.9\,$\pm$\,0.2 & 81.9\,$\pm$\,0.2 \\
 & Full ReCalMatch        & \textbf{54.2\,$\pm$\,0.4} & \textbf{78.2\,$\pm$\,0.3} & \textbf{82.3\,$\pm$\,0.2} \\
\bottomrule
\end{tabular}
\end{table}

\subsection{Aspect and Gating Sensitivity}
Figure~\ref{fig:aspect_config} reports the effect of including different semantic aspects. Accuracy increases monotonically as aspects are added, with the part aspect being the strongest single aspect and the full color+shape+part configuration consistently the best. This complementarity is what makes the cross-aspect disagreement $d_j$ informative: redundant aspects would covary, collapsing $d_j$. Figure~\ref{fig:zeta_sensitivity} shows that ReCalMatch is robust to the disagreement scale $\zeta$, following an inverted-U that peaks at $\zeta=2$ and stays stable within $\zeta\in[1,3]$; $\zeta=0$ disables the gating and recovers the configuration without disagreement.

\subsection{Sensitivity, Stability, and Efficiency}

Table~\ref{tab:sensitivity} summarizes hyperparameter sensitivity and stability. ReCalMatch is robust to the loss weights $(\lambda_s,\lambda_m)$, with the default values at the peak. For the confidence and entropy exponents $(\alpha,\beta)$, using either factor alone improves over the unweighted variant, and combining both is strongest; overly large exponents slightly suppress informative samples. Temperatures $\tau_g$ and $\tau_c$ both peak at $0.07$ and remain stable within $[0.05,0.10]$. Multi-seed results confirm that the gains, including the disagreement-gating increment, are stable across random splits.

The semantic bank is encoded offline once per dataset, so the online calibration itself consists mainly of cosine similarities and a closed-form disagreement statistic. ReCalMatch uses $17.6$ GFLOPs at $0.120$ s/step, close to SemiVisBooster in FLOPs ($17.5$ GFLOPs) but substantially faster than its $0.198$ s/step runtime, while remaining more expensive than the visual-only FreeMatch baseline ($4.2$ GFLOPs, $0.060$ s/step). This trade-off is most favorable in the low-label regimes where ReCalMatch provides its largest accuracy gains.

\vspace{-0.4em}
\subsection{Semantic Bank Source}
\begin{wraptable}{r}{0.43\linewidth}\centering\vspace{-0.8em}\small\caption{Effect of semantic-bank source on CUB-200-2011 under the 1\% labeled-data setting.}\label{tab:semantic_source}\begin{tabular}{l|c}\toprule\textbf{Semantic-bank source} & \textbf{Top-1 (\%)} \\\midrule Class names only & 50.9 \\ LLM-generated aspects & 53.6 \\ Curated metadata aspects & 54.2 \\ LLM-generated + metadata & \textbf{54.8} \\\bottomrule\end{tabular}\vspace{-0.8em}\end{wraptable}Since ReCalMatch uses semantic prototypes as external calibration evidence, we further study whether its performance depends on curated dataset metadata. Table~\ref{tab:semantic_source} compares different sources for constructing the semantic bank on CUB-200-2011 under the 1\% labeled-data setting. Using only class names already provides a strong improvement over the visual-only baseline, while an LLM-generated semantic bank without curated attributes achieves 53.6\% accuracy, close to the default metadata-based version. Combining LLM-generated descriptions with available metadata further improves performance to 54.8\% accuracy. These results suggest that ReCalMatch does not rely strictly on manually curated attributes; such metadata mainly provides a cleaner instantiation of the general semantic-bank interface.

\vspace{-2em}
\section{Conclusion}
\label{sec:conclusion}
\vspace{-1em}

We presented ReCalMatch, a reliability-calibrated semantic framework for semi-supervised fine-grained recognition. Rather than relying solely on visual confidence, ReCalMatch introduces multi-aspect semantic evidence as a reliability signal. It constructs class-conditioned semantic prototypes, measures visual--semantic agreement for each pseudo-label, and suppresses unreliable samples through cross-aspect disagreement-aware calibration. Together with semantic consistency and prototype margin regularization, this design reduces overconfident pseudo-label errors and improves pseudo-label learning in fine-grained SSL. ReCalMatch depends on the quality and domain relevance of its semantic bank; when visual appearance genuinely contradicts the prior, such as atypical coloration, high disagreement may over-suppress a correct pseudo-label. Domain-specific or LLM-generated aspects mitigate reliance on curated attributes, but extending semantic reliability calibration to domains with substantially different cues and to dense or open-vocabulary prediction remains future work.

\bibliography{egbib}

@article{wei2021fine,
  title={Fine-grained image analysis with deep learning: A survey},
  author={Wei, Xiu-Shen and Song, Yi-Zhe and Mac Aodha, Oisin and Wu, Jianxin and Peng, Yuxin and Tang, Jinhui and Yang, Jian and Belongie, Serge},
  journal={IEEE transactions on pattern analysis and machine intelligence},
  volume={44},
  number={12},
  pages={8927--8948},
  year={2021},
  publisher={IEEE}
}

@inproceedings{zheng2022simmatch,
  title={Simmatch: Semi-supervised learning with similarity matching},
  author={Zheng, Mingkai and You, Shan and Huang, Lang and Wang, Fei and Qian, Chen and Xu, Chang},
  booktitle={2022 IEEE/CVF Conference on Computer Vision and Pattern Recognition (CVPR)},
  pages={14451--14461},
  year={2022},
  organization={IEEE}
}

@inproceedings{zhu2022dual,
  title={Dual cross-attention learning for fine-grained visual categorization and object re-identification},
  author={Zhu, Haowei and Ke, Wenjing and Li, Dong and Liu, Ji and Tian, Lu and Shan, Yi},
  booktitle={Proceedings of the IEEE/CVF conference on computer vision and pattern recognition},
  pages={4692--4702},
  year={2022}
}

@article{dosovitskiy2020image,
  title={An image is worth 16x16 words: Transformers for image recognition at scale},
  author={Dosovitskiy, Alexey},
  journal={arXiv preprint arXiv:2010.11929},
  year={2020}
}

@inproceedings{he2022transfg,
  title={Transfg: A transformer architecture for fine-grained recognition},
  author={He, Ju and Chen, Jie-Neng and Liu, Shuai and Kortylewski, Adam and Yang, Cheng and Bai, Yutong and Wang, Changhu},
  booktitle={Proceedings of the AAAI conference on artificial intelligence},
  volume={36},
  number={1},
  pages={852--860},
  year={2022}
}

@inproceedings{radford2021learning,
  title={Learning transferable visual models from natural language supervision},
  author={Radford, Alec and Kim, Jong Wook and Hallacy, Chris and Ramesh, Aditya and Goh, Gabriel and Agarwal, Sandhini and Sastry, Girish and Askell, Amanda and Mishkin, Pamela and Clark, Jack and others},
  booktitle={International conference on machine learning},
  pages={8748--8763},
  year={2021},
  organization={PmLR}
}

@inproceedings{krause20133d,
  title={3d object representations for fine-grained categorization},
  author={Krause, Jonathan and Stark, Michael and Deng, Jia and Fei-Fei, Li},
  booktitle={Proceedings of the IEEE international conference on computer vision workshops},
  pages={554--561},
  year={2013}
}

@inproceedings{fu2017look,
  title={Look closer to see better: Recurrent attention convolutional neural network for fine-grained image recognition},
  author={Fu, Jianlong and Zheng, Heliang and Mei, Tao},
  booktitle={Proceedings of the IEEE conference on computer vision and pattern recognition},
  pages={4438--4446},
  year={2017}
}

@inproceedings{zhang2014part,
  title={Part-based R-CNNs for fine-grained category detection},
  author={Zhang, Ning and Donahue, Jeff and Girshick, Ross and Darrell, Trevor},
  booktitle={European conference on computer vision},
  pages={834--849},
  year={2014},
  organization={Springer}
}

@inproceedings{zheng2017learning,
  title={Learning multi-attention convolutional neural network for fine-grained image recognition},
  author={Zheng, Heliang and Fu, Jianlong and Mei, Tao and Luo, Jiebo},
  booktitle={Proceedings of the IEEE international conference on computer vision},
  pages={5209--5217},
  year={2017}
}

@inproceedings{lin2015bilinear,
  title={Bilinear CNN models for fine-grained visual recognition},
  author={Lin, Tsung-Yu and RoyChowdhury, Aruni and Maji, Subhransu},
  booktitle={Proceedings of the IEEE international conference on computer vision},
  pages={1449--1457},
  year={2015}
}

@inproceedings{gao2016compact,
  title={Compact bilinear pooling},
  author={Gao, Yang and Beijbom, Oscar and Zhang, Ning and Darrell, Trevor},
  booktitle={Proceedings of the IEEE conference on computer vision and pattern recognition},
  pages={317--326},
  year={2016}
}

@article{wang2021feature,
  title={Feature fusion vision transformer for fine-grained visual categorization},
  author={Wang, Jun and Yu, Xiaohan and Gao, Yongsheng},
  journal={arXiv preprint arXiv:2107.02341},
  year={2021}
}

@inproceedings{farhadi2009describing,
  title={Describing objects by their attributes},
  author={Farhadi, Ali and Endres, Ian and Hoiem, Derek and Forsyth, David},
  booktitle={2009 IEEE conference on computer vision and pattern recognition},
  pages={1778--1785},
  year={2009},
  organization={IEEE}
}

@article{lampert2013attribute,
  title={Attribute-based classification for zero-shot visual object categorization},
  author={Lampert, Christoph H and Nickisch, Hannes and Harmeling, Stefan},
  journal={IEEE transactions on pattern analysis and machine intelligence},
  volume={36},
  number={3},
  pages={453--465},
  year={2013},
  publisher={IEEE}
}

@article{zhou2022learning,
  title={Learning to prompt for vision-language models},
  author={Zhou, Kaiyang and Yang, Jingkang and Loy, Chen Change and Liu, Ziwei},
  journal={International Journal of Computer Vision},
  volume={130},
  number={9},
  pages={2337--2348},
  year={2022},
  publisher={Springer}
}

@inproceedings{mao2023doubly,
  title={Doubly right object recognition: A why prompt for visual rationales},
  author={Mao, Chengzhi and Teotia, Revant and Sundar, Amrutha and Menon, Sachit and Yang, Junfeng and Wang, Xin and Vondrick, Carl},
  booktitle={Proceedings of the IEEE/CVF Conference on Computer Vision and Pattern Recognition},
  pages={2722--2732},
  year={2023}
}

@article{A2,
  title={Realistic evaluation of deep semi-supervised learning algorithms},
  author={Oliver, Avital and Odena, Augustus and Raffel, Colin A and Cubuk, Ekin Dogus and Goodfellow, Ian},
  journal={Advances in neural information processing systems},
  volume={31},
  year={2018}
}

@article{A4,
  title={Flexmatch: Boosting semi-supervised learning with curriculum pseudo labeling},
  author={Zhang, Bowen and Wang, Yidong and Hou, Wenxin and Wu, Hao and Wang, Jindong and Okumura, Manabu and Shinozaki, Takahiro},
  journal={Advances in neural information processing systems},
  volume={34},
  pages={18408--18419},
  year={2021}
}

@article{A5,
  title={Selfmatch: Combining contrastive self-supervision and consistency for semi-supervised learning},
  author={Kim, Byoungjip and Choo, Jinho and Kwon, Yeong-Dae and Joe, Seongho and Min, Seungjai and Gwon, Youngjune},
  journal={arXiv preprint arXiv:2101.06480},
  year={2021}
}

@inproceedings{A7,
  title={Pseudo-labeling and confirmation bias in deep semi-supervised learning},
  author={Arazo, Eric and Ortego, Diego and Albert, Paul and O’Connor, Noel E and McGuinness, Kevin},
  booktitle={2020 International joint conference on neural networks (IJCNN)},
  pages={1--8},
  year={2020},
  organization={IEEE}
}

@inproceedings{A11,
  title={The inaturalist species classification and detection dataset},
  author={Van Horn, Grant and Mac Aodha, Oisin and Song, Yang and Cui, Yin and Sun, Chen and Shepard, Alex and Adam, Hartwig and Perona, Pietro and Belongie, Serge},
  booktitle={Proceedings of the IEEE conference on computer vision and pattern recognition},
  pages={8769--8778},
  year={2018}
}

@inproceedings{A12,
  title={Building a bird recognition app and large scale dataset with citizen scientists: The fine print in fine-grained dataset collection},
  author={Van Horn, Grant and Branson, Steve and Farrell, Ryan and Haber, Scott and Barry, Jessie and Ipeirotis, Panos and Perona, Pietro and Belongie, Serge},
  booktitle={Proceedings of the IEEE conference on computer vision and pattern recognition},
  pages={595--604},
  year={2015}
}

@techreport{A13,
  title={The caltech-ucsd birds-200-2011 dataset},
  author={Wah, Catherine and Branson, Steve and Welinder, Peter and Perona, Pietro and Belongie, Serge},
  year={2011},
  publisher={California Institute of Technology}
}

@inproceedings{standforddogs,
author = "Aditya Khosla and Nityananda Jayadevaprakash and Bangpeng Yao and Li Fei-Fei",
title = "Novel Dataset for Fine-Grained Image Categorization",
booktitle = "First Workshop on Fine-Grained Visual Categorization, IEEE Conference on Computer Vision and Pattern Recognition",
year = "2011",
month = "June",
address = "Colorado Springs, CO",
}

@article{A18,
  title={Semi-supervised learning by entropy minimization},
  author={Grandvalet, Yves and Bengio, Yoshua},
  journal={Advances in neural information processing systems},
  volume={17},
  year={2004}
}

@article{A20,
  title={Semi-supervised vision transformers at scale},
  author={Cai, Zhaowei and Ravichandran, Avinash and Favaro, Paolo and Wang, Manchen and Modolo, Davide and Bhotika, Rahul and Tu, Zhuowen and Soatto, Stefano},
  journal={Advances in Neural Information Processing Systems},
  volume={35},
  pages={25697--25710},
  year={2022}
}

@article{A28,
  title={Weakly supervised bilinear attention network for fine-grained visual classification},
  author={Hu, Tao and Xu, Jizheng and Huang, Cong and Qi, Honggang and Huang, Qingming and Lu, Yan},
  journal={arXiv preprint arXiv:1808.02152},
  year={2018}
}

@article{A31,
  title={Remixmatch: Semi-supervised learning with distribution alignment and augmentation anchoring},
  author={Berthelot, David and Carlini, Nicholas and Cubuk, Ekin D and Kurakin, Alex and Sohn, Kihyuk and Zhang, Han and Raffel, Colin},
  journal={arXiv preprint arXiv:1911.09785},
  year={2019}
}

@inproceedings{A32,
  title={Pepl: Precision-enhanced pseudo-labeling for fine-grained image classification in semi-supervised learning},
  author={Tian, Bowen and Lai, Songning and Li, Lujundong and Shuai, Zhihao and Guan, Runwei and Wu, Tian and Yue, Yutao},
  booktitle={ICASSP 2025-2025 IEEE International Conference on Acoustics, Speech and Signal Processing (ICASSP)},
  pages={1--5},
  year={2025},
  organization={IEEE}
}

@inproceedings{A33,
  title={Roll with the punches: expansion and shrinkage of soft label selection for semi-supervised fine-grained learning},
  author={Duan, Yue and Zhao, Zhen and Qi, Lei and Zhou, Luping and Wang, Lei and Shi, Yinghuan},
  booktitle={Proceedings of the AAAI Conference on Artificial Intelligence},
  volume={38},
  pages={11829--11837},
  year={2024}
}

@article{A42,
  title={Softmatch: Addressing the quantity-quality trade-off in semi-supervised learning},
  author={Chen, Hao and Tao, Ran and Fan, Yue and Wang, Yidong and Wang, Jindong and Schiele, Bernt and Xie, Xing and Raj, Bhiksha and Savvides, Marios},
  journal={arXiv preprint arXiv:2301.10921},
  year={2023}
}

@inproceedings{rao2021counterfactual,
  title={Counterfactual attention learning for fine-grained visual categorization and re-identification},
  author={Rao, Yongming and Chen, Guangyi and Lu, Jiwen and Zhou, Jie},
  booktitle={Proceedings of the IEEE/CVF international conference on computer vision},
  pages={1025--1034},
  year={2021}
}

@inproceedings{zhang2022free,
  title={A free lunch from vit: Adaptive attention multi-scale fusion transformer for fine-grained visual recognition},
  author={Zhang, Yuan and Cao, Jian and Zhang, Ling and Liu, Xiangcheng and Wang, Zhiyi and Ling, Feng and Chen, Weiqian},
  booktitle={ICASSP 2022-2022 IEEE International conference on acoustics, speech and signal processing (ICASSP)},
  pages={3234--3238},
  year={2022},
  organization={IEEE}
}

@article{xu2023fine,
  title={Fine-grained visual classification via internal ensemble learning transformer},
  author={Xu, Qin and Wang, Jiahui and Jiang, Bo and Luo, Bin},
  journal={IEEE Transactions on Multimedia},
  volume={25},
  pages={9015--9028},
  year={2023},
  publisher={IEEE}
}

@inproceedings{hu2021rams,
  title={Rams-trans: Recurrent attention multi-scale transformer for fine-grained image recognition},
  author={Hu, Yunqing and Jin, Xuan and Zhang, Yin and Hong, Haiwen and Zhang, Jingfeng and He, Yuan and Xue, Hui},
  booktitle={Proceedings of the 29th ACM international conference on multimedia},
  pages={4239--4248},
  year={2021}
}

@inproceedings{sun2022sim,
  title={Sim-trans: Structure information modeling transformer for fine-grained visual categorization},
  author={Sun, Hongbo and He, Xiangteng and Peng, Yuxin},
  booktitle={Proceedings of the 30th ACM international conference on multimedia},
  pages={5853--5861},
  year={2022}
}

@inproceedings{zhang2025semivisbooster,
  title={SemiVisBooster: Boosting Semi-Supervised Learning for Fine-Grained Classification through Pseudo-Label Semantic Guidance},
  author={Zhang, Wenjin and Li, Xinyu and Gao, Chenyang and Marsic, Ivan},
  booktitle={Proceedings of the IEEE/CVF International Conference on Computer Vision},
  pages={1195--1204},
  year={2025}
}

@article{berthelot2021adamatch,
  title={Adamatch: A unified approach to semi-supervised learning and domain adaptation},
  author={Berthelot, David and Roelofs, Rebecca and Sohn, Kihyuk and Carlini, Nicholas and Kurakin, Alex},
  journal={arXiv preprint arXiv:2106.04732},
  year={2021}
}

@inproceedings{xu2021dash,
  title={Dash: Semi-supervised learning with dynamic thresholding},
  author={Xu, Yi and Shang, Lei and Ye, Jinxing and Qian, Qi and Li, Yu-Feng and Sun, Baigui and Li, Hao and Jin, Rong},
  booktitle={International conference on machine learning},
  pages={11525--11536},
  year={2021},
  organization={PMLR}
}

@inproceedings{usb2022,
  doi = {10.48550/ARXIV.2208.07204},
  url = {https://arxiv.org/abs/2208.07204},
  author = {Wang, Yidong and others},
  title = {USB: A Unified Semi-supervised Learning Benchmark for Classification},
  booktitle = {Thirty-sixth Conference on Neural Information Processing Systems Datasets and Benchmarks Track},
  year = {2022}
}

@inproceedings{he2025semi,
  title={Semi-ViM: Bidirectional State Space Model for Mitigating Label Imbalance in Semi-Supervised Learning},
  author={He, Hongyang and Xie, Hongyang and You, Haochen and Sanchez, Victor},
  booktitle={Proceedings of the IEEE/CVF International Conference on Computer Vision},
  pages={765--774},
  year={2025}
}

@article{berthelot2019mixmatch,
  title={Mixmatch: A holistic approach to semi-supervised learning},
  author={Berthelot, David and Carlini, Nicholas and Goodfellow, Ian and Papernot, Nicolas and Oliver, Avital and Raffel, Colin A},
  journal={Advances in neural information processing systems},
  volume={32},
  year={2019}
}

@article{sohn2020fixmatch,
  title={Fixmatch: Simplifying semi-supervised learning with consistency and confidence},
  author={Sohn, Kihyuk and others},
  journal={Advances in neural information processing systems},
  volume={33},
  pages={596--608},
  year={2020}
}

@article{wang2022freematch,
  title={Freematch: Self-adaptive thresholding for semi-supervised learning},
  author={Wang, Yidong and Chen, Hao and Heng, Qiang and Hou, Wenxin and Fan, Yue and Wu, Zhen and Wang, Jindong and Savvides, Marios and Shinozaki, Takahiro and Raj, Bhiksha and others},
  journal={arXiv preprint arXiv:2205.07246},
  year={2022}
}

@inproceedings{su2021realistic,
  title={A realistic evaluation of semi-supervised learning for fine-grained classification},
  author={Su, Jong-Chyi and Cheng, Zezhou and Maji, Subhransu},
  booktitle={Proceedings of the IEEE/CVF conference on computer vision and pattern recognition},
  pages={12966--12975},
  year={2021}
}

@inproceedings{lee2013pseudo,
  title={Pseudo-label: The simple and efficient semi-supervised learning method for deep neural networks},
  author={Lee, Dong-Hyun and others},
  booktitle={Workshop on challenges in representation learning, ICML},
  volume={3},
  number={2},
  pages={896},
  year={2013},
  organization={Atlanta}
}

@article{laine2016temporal,
  title={Temporal ensembling for semi-supervised learning},
  author={Laine, Samuli and Aila, Timo},
  journal={arXiv preprint arXiv:1610.02242},
  year={2016}
}

@article{tarvainen2017mean,
  title={Mean teachers are better role models: Weight-averaged consistency targets improve semi-supervised deep learning results},
  author={Tarvainen, Antti and Valpola, Harri},
  journal={Advances in neural information processing systems},
  volume={30},
  year={2017}
}

@article{mugnai2022fine,
  title={Fine-grained adversarial semi-supervised learning},
  author={Mugnai, Daniele and Pernici, Federico and Turchini, Francesco and Del Bimbo, Alberto},
  journal={ACM Transactions on Multimedia Computing, Communications, and Applications (TOMM)},
  volume={18},
  number={1s},
  pages={1--19},
  year={2022},
  publisher={ACM New York, NY}
}

@article{nartey2019semi,
  title={Semi-supervised learning for fine-grained classification with self-training},
  author={Nartey, Obed Tettey and Yang, Guowu and Wu, Jinzhao and Asare, Sarpong Kwadwo},
  journal={Ieee Access},
  volume={8},
  pages={2109--2121},
  year={2019},
  publisher={IEEE}
}

@article{xie2020unsupervised,
  title={Unsupervised data augmentation for consistency training},
  author={Xie, Qizhe and Dai, Zihang and Hovy, Eduard and Luong, Thang and Le, Quoc},
  journal={Advances in neural information processing systems},
  volume={33},
  pages={6256--6268},
  year={2020}
}

@inproceedings{linpart,
  title={Part-level Semantic-guided Contrastive Learning for Fine-grained Visual Classification},
  author={Lin, Zhijian and Han, Hong},
  booktitle={The Fourteenth International Conference on Learning Representations}
}

@article{fang2025low,
  title={A low functional redundancy-based network slimming method for accelerating deep neural networks},
  author={Fang, Zheng and Yin, Bo},
  journal={Alexandria Engineering Journal},
  volume={119},
  pages={437--450},
  year={2025},
  publisher={Elsevier}
}

@article{fan2024context,
  title={Context-Based Semantic-Aware Alignment for Semi-Supervised Multi-Label Learning},
  author={Fan, Heng-Bo and Xie, Ming-Kun and Xiao, Jia-Hao and Huang, Sheng-Jun},
  journal={arXiv preprint arXiv:2412.18842},
  year={2024}
}

@article{liu2024multi,
  title={Multi-grained cross-modal alignment for learning open-vocabulary semantic segmentation from text supervision},
  author={Liu, Yajie and Ge, Pu and Liu, Qingjie and Huang, Di},
  journal={arXiv preprint arXiv:2403.03707},
  year={2024}
}

@inproceedings{A46,
  title={Research on surface defect detection method of metal workpiece based on machine learning},
  author={He, Hongyang and Yuan, Mingang and Liu, Xiushan},
  booktitle={2021 6th international conference on Intelligent Computing and Signal Processing (ICSP)},
  pages={881--884},
  year={2021},
  organization={IEEE}
}

@inproceedings{A47,
  title={Research on the application of electronic technology of internet of things in smart city},
  author={He, Hongyang},
  booktitle={2020 International Conference on Intelligent Transportation, Big Data \& Smart City (ICITBS)},
  pages={454--457},
  year={2020},
  organization={IEEE}
}

@inproceedings{A48,
  title={TrustMatch: mitigating pseudo-label bias in semi-supervised learning with trust-aware refinement},
  author={He, Hongyang and Hong, Yundi},
  booktitle={2025 IEEE/CVF International Conference on Computer Vision Workshops (ICCVW)},
  pages={605--614},
  year={2025},
  organization={IEEE}
}

@inproceedings{A49,
  title={Research on pedestrian tracking algorithm based on deep learning},
  author={He, Hongyang and Yan, Ziran and Geng, Zichao and Liu, Xiushan},
  booktitle={2021 International Conference on Computer Information Science and Artificial Intelligence (CISAI)},
  pages={487--490},
  year={2021},
  organization={IEEE}
}

@inproceedings{A50,
  title={Semi-vim: bidirectional state space model for mitigating label imbalance in semi-supervised learning},
  author={He, Hongyang and Xie, Hongyang and You, Haochen and Sanchez, Victor},
  booktitle={2025 IEEE/CVF International Conference on Computer Vision (ICCV)},
  pages={765--774},
  year={2025},
  organization={IEEE}
}

@article{A51,
  title={Traditional Machine Learning Models for Building Energy Performance Prediction: A Comparative Research},
  author={Wu, Zeyu and He, Hongyang},
  journal={Machine Learning Research},
  volume={8},
  number={1},
  pages={1--8},
  year={2023}
}

@inproceedings{A52,
  title={Token-aware representation augmentation for fine-grained semi-supervised learning},
  author={He, Hongyang and Zhong, Yan and Song, Xinyuan and Liu, Daizong and Sanchez, Victor},
  booktitle={The Third Conference on Parsimony and Learning (Proceedings Track)},
  year={2026}
}

@article{A53,
  title={TRiCo: Triadic Game-Theoretic Co-Training for Robust Semi-Supervised Learning},
  author={He, Hongyang and Song, Xinyuan and He, Yangfan and Zhang, Zeyu and Li, Yanshu and You, Haochen and Sun, Lifan and Zhang, Wenqiao},
  journal={Advances in Neural Information Processing Systems},
  volume={38},
  pages={87545--87570},
  year={2026}
}

@inproceedings{A54,
  title={4S-Classifier: Empowering Conservation through Semi-Supervised Learning for Rare and Endangered Species},
  author={He, Hongyang and Xie, Hongyang and Shen, Guodong and Fu, Boyang and You, Haochen and Sanchez, Victor},
  booktitle={2025 International Joint Conference on Neural Networks (IJCNN)},
  pages={1--10},
  year={2025},
  organization={IEEE}
}

@article{A55,
  title={Revisiting Chain-of-Thought Reasoning under Limited Supervision: Semi-supervised Chain-of-Thought Learning},
  author={He, Hongyang and Liu, Jiuming and Sanchez, Victor},
  journal={arXiv preprint arXiv:2607.01511},
  year={2026}
}

@article{A56,
  title={Semi-Supervised Vision-Language-Action Model},
  author={He, Hongyang and Liu, Jiuming and Sanchez, Victor},
  journal={arXiv preprint arXiv:2606.21493},
  year={2026}
}

@inproceedings{A57,
  title={Newton-coupled Dual-Teacher Semi-supervised Learning Framework},
  author={He, Hongyang and Song, Xinyuan and Zhong, Yan and Liu, Daizong and Liu, Xuanyu and Sanchez, Victor},
  booktitle={Forty-third International Conference on Machine Learning},
  year={2026}
}

@inproceedings{A58,
  title={Beyond Data Augmentation: Energy-Based Kuramoto Neurons for Semi-Supervised Learning},
  author={Hong, Yundi and Li, Ao and Song, Xinyuan and Zhong, Yan and He, Hongyang and Sanchez, Victor},
  booktitle={Proceedings of the IEEE/CVF Conference on Computer Vision and Pattern Recognition},
  pages={2913--2922},
  year={2026}
}

@article{A59,
  title={YOLO-LRDD: A lightweight method for road damage detection based on improved YOLOv5s},
  author={Wan, Fang and Sun, Chen and He, Hongyang and Lei, Guangbo and Xu, Li and Xiao, Teng},
  journal={EURASIP Journal on Advances in Signal Processing},
  volume={2022},
  number={1},
  pages={98},
  year={2022},
  publisher={Springer}
}

@article{A60,
  title={Triple Expert Learning from Noisy Labels for Semi-Supervised Vision Foundation Model Adaptation},
  author={Liu, Xuanyu and Fang, Zheng and He, Hongyang and Hong, Yundi and Liu, Daizong},
  journal={arXiv preprint arXiv:2608.09052},
  year={2026}
}

@article{A63,
  title={Geometric Regularization for Long-Tailed Semi-Supervised Learning via Gaussian Feature Bridges},
  author={He, Hongyang and Song, Xinyuan and Zhong, Yan and Liu, Daizong and Li, Yanbin and He, Yang-fan and Zhang, Wenqiao},
  journal={arXiv preprint arXiv:2608.20710},
  year={2026}
}
\end{document}